\documentclass{amia}
\usepackage{graphicx}
\usepackage[labelfont=bf]{caption}
\usepackage[superscript,nomove]{cite}
\usepackage[dvipsnames]{xcolor}
\usepackage[sort,comma,numbers,super]{natbib}
\usepackage{multirow}
\usepackage{url}
\usepackage{adjustbox}
\usepackage{tabularx}
\usepackage{soul}
\usepackage{ulem} % Add this line to your LaTeX preamble
\usepackage{tabularx}
\usepackage{hyperref}
\usepackage{listings}
\usepackage{amsmath}
\usepackage{needspace}
\usepackage{enumitem}

\usepackage{booktabs}
\usepackage{multirow}
\usepackage{makecell}
\usepackage{array}
\usepackage{pifont}
\usepackage{bm}
\usepackage{relsize}
\usepackage{amsmath}
\usepackage{subcaption}
\usepackage{sidecap}
\usepackage{fancyhdr}
\usepackage{csquotes}
\newcommand\EatDot[1]{}

\definecolor{myblue}{HTML}{4E79A7}
\definecolor{mygreen}{HTML}{59A14F}
\definecolor{myred}{HTML}{E15759}

\begin{document}

\title{Text Classification of Cancer Clinical Trial Eligibility Criteria }

\author{Yumeng Yang, MS$^1$, Soumya Jayaraj, BAT$^1$, Ethan Ludmir, MD$^2$ $^3$, Kirk Roberts, PhD$^1$} 

\institutes{
    $^1$School of Biomedical Informatics\\ The University of Texas Health Science Center at Houston, Houston, TX, USA \\
    $^2$Department of Radiation Oncology\\ The University of Texas MD Anderson Cancer Center, Houston, TX, USA\\
    $^3$Department of Biostatistics \\The University of Texas MD Anderson Cancer Center, Houston, TX, USA
\\
}
\maketitle

%\vspace{-5pt}
\section*{Abstract}
\label{Abstarct}
%\vspace{-5pt}
%Our study aims to develop automated classification models for eligibility criteria for cancer trials from ClinicalTrials.gov to facilitate the patient-trial matching process. We started from 764 annotated trials, using keywords matching to extract criteria that specifically convey certain criteria requirement. Alone with training data with five publicly available domain specific-models, we also pretrained our own model by using all avilable documents from ClinicalTrials.gov as of February 15t, then tested it on our classification task, the best f1 score for all criteria, prior malignancy, HIV, HBV, HCV, psycharistic illness, combination of alcohol, drug and substance abuse(SubDrugAlc), and autoimmune is over 0.95. Our result shows that it is feasible to build a classification model that automatically identify high level and frequent asked eligibility criteria for cancer clinical trials. Furthermore, our pretrained model outperforms other models in some criteria. Future work focused on studying eligibility criteria for clinical trials can greatly benefit from our model.
Automatic identification of clinical trials for which a patient is eligible is complicated by the fact that trial eligibility is stated in natural language. A potential solution to this problem is to employ text classification methods for common types of eligibility criteria. In this study, we focus on seven common exclusion criteria in cancer trials: prior malignancy, human immunodeficiency virus, hepatitis B, hepatitis C, psychiatric illness, drug/substance abuse, and autoimmune illness. Our dataset consists of 764 phase III cancer trials with these exclusions annotated at the trial level. We experiment with common transformer models as well as a new pre-trained clinical trial BERT model. Our results demonstrate the feasibility of automatically classifying common exclusion criteria. Additionally, we demonstrate the value of a pre-trained language model specifically for clinical trials, which yields the highest average performance across all criteria.

%\vspace{-5pt}
\section{Introduction}
\label{Introduction}
%\vspace{-5pt}

Cancer has a high morbidity and mortality rate and threatens millions of people’s lives. According to the American Cancer Society report, there will be a total of 1.9 million new cancer cases alone with more than 600,000 deaths in the US in 2022\cite{acs2022}. Clinical trials have always been recognized as significant for cancer treatment and anticancer drug development\cite{cox2003patients}, but trial recruitment is still problematic, with incomplete accrual as the leading reason for non-informative trial closure/failure\cite{cox2003patients,kadam2016challenges,ross1999barriers}. Barriers to trial accrual are multi-factorial, including patient and provider concerns and biases, as well as the availability of trials across different centers\cite{jones2007identifying,jenkins2010attitudes,mills2006barriers}. Critically, matching potentially eligible patients to relevant clinical trials is a key barrier to clinical trial accrual, as patients seeking trial options aim to identify candidate trials for which they are eligible. Eligibility criteria for clinical trials define which patients are eligible for a given study; however, trial eligibility criteria are often written in jargon and are often difficult for lay audiences to interpret, understand, and apply\cite{kang2015initial}. 

Eligibility criteria are a key component of any clinical trial or study protocol, as they define the requirements for participation, and indeed define the study population.  Conventionally, eligibility criteria include both inclusion and exclusion criteria\cite{usnll2022clinicaltrials}. Eligibility criteria are generally listed in clinical trial registries as well; ClinicalTrials.gov is a public database that provides information on registered human trials. It is managed by the United States National Library of Medicine (NLM)\cite{nlm2013clinicaltrials}, and federal mandates require that human trials in the US are registered with trial information in ClinicalTrials.gov\cite{clinicaltrials_gov}. As of February 2023, approximately half a million studies were registered in ClinicalTrials.gov, making it the largest available trial registry worldwide. Such a comprehensive database brings enormous potential for research, including identifying high-level trends in trial structure, evaluation of disparities\cite{grant2020racial,corrigan2022inclusion}, and developing tools to help facilitate trial recruitment.

Identifying eligible patients for these trials can be a time-consuming and challenging task, partly due to the non-standardized format of eligibility criteria. As a result, patients and clinicians may struggle to identify relevant trials, leading to potential delays in the accrual process. \cite{bhattacharya2013analysis}. This underscores the urgent need for automated tools, such as text classification, to streamline and enhance the recruitment process. By developing an automatic classifier to identify key eligibility criteria from free-text records, we can potentially address the unmet needs of patients and clinicians by facilitating faster and more accurate identification of relevant clinical trials. Natural Language Processing (NLP) aims to enable machine understanding of human language. In the clinical field, NLP has a wide range of applications, including name entity recognition\cite{zhao2004named,ji2019hybrid,bhatia2019comprehend}, text mining\cite{eom2004pubminer,huang2008genclip,bucur2014supporting}, and text classification\cite{uzuner2008identifying,nii2017nursing,yao2019clinical}. NLP models can help extract and structure information from text data, ranging from patients' clinical notes to trial eligibility criteria. Our project aims to develop classifiers that can automatically identify key exclusion criteria from eligibility descriptions listed on ClinicalTrials.gov. Such classification tools have the potential to streamline the accrual process for both clinicians and patients. For our study, we chose seven key eligibility criteria and used five state-of-the-art domain-specific language models trained on our data. Additionally, we pre-trained our own model based on ClinicalBERT by using approximately half a million eligibility criteria sections derived from ClinicalTrials.gov.

%\vspace{-5pt}
\section{Related work}
\label{Related work}
%\vspace{-5pt}
% add more about NLP Classification models

% can be divdied into ``texting mining tool'', ``information extraction'', ``avilable corpus'' -- similar to previous literature review
Numerous prior works have focused on text mining of eligibility criteria for clinical trials for the purpose of streamlining the recruitment process for both patients and clinicians. Criteria2Query is an NLP tool aiming to convert free-text eligibility criteria for a clinical trial to a structured query to aid in the identification of eligible patients from clinical data\cite{yuan2019criteria2query}. DQueST is a dynamic questionnaire to help find eligible clinical trials by asking trial criteria-related questions\cite{liu2019dquest}. There also exists many models to automatically structure eligibility criteria for clinical trials\cite{luo2011extracting} and EHR data\cite{jonnalagadda2017text}. RuleEd is a web-based tool to revise and refine free text eligibility criteria\cite{olasov2006ruleed}. EXTRACTS is a search tool that allows users to customize criteria and weight each criterion for potential trials\cite{miotto2013etacts}.

Another aspect of emphasis is information extraction combined with machine learning techniques to match terms from eligibility criteria and patient records. Some tools used regular expression and machine learning models to identify certain criteria for cancer trials\cite{zhang2017automated} and to satisfy specific department needs\cite{ni2015automated}. Another tool was developed to identify eligible trials using key terms and patterns matching\cite{petkov2013automated}. Many other tools were developed based upon EHR data to further identify and match eligible patients from their medical records for specific diseases, such as cancer and Alzheimer's disease\cite{kirshner2021automated,tissot2020natural,cai2021improving} with clinical trial criteria. 

Along with text mining and information extraction, some works aim to create a knowledge base of common eligibility criteria related to annotated corpus and build the knowledge base. EliIE\cite{kang2017eliie} contains 230 trials from various phases with a focus on Alzheimer’s disease specifically, Chia\cite{kury2020chia} contains 1000 phase IV trials covering all diseases, and The Leaf Clinical Trials Corpus\cite{dobbins2022leaf} contains 1006 trials cross all phases and diseases. A lexicon base for breast cancer clinical trial eligibility criteria was created to identify concepts related to eligibility\cite{jung2021building}. This study shows that a specified lexicon can improve the accuracy of subjects in clinical trial eligibility criteria analysis. Another knowledge base used a hierarchical taxonomy to classify criteria into multiple categories, including disease, intervention, and condition.\cite{liu2021knowledge}. 

%\vspace{-5pt}
\section{Method}
%\vspace{-5pt}
Our dataset is based on PROTECTOR1, a manually annotated database of 764 Phase III cancer trials collected from ClinicalTrials.gov covering the years 2000-2017.
PROTECTOR1 has been used to analyze many aspects of cancer trials, including evaluating the relationship between sponsorship types with trial accural\cite{pasalic2020association}, measure impacts and trial-related factors for these excluded patients with brain metastases in cancer trials\cite{patel2020exclusion}, and ascertain the transparency regarding cancer trial results reports\cite{patel2021transparency}. % TODO cite some of these: https://scholar.google.com/scholar?hl=en&as_sdt=0%2C44&q=author%3A%22e+ludmir%22+764&btnG=
The database was originally developed for manual analysis, which resulted in several important design decisions that will be described in detail later in Section~\ref{section:keyword_filtering} (notably, criteria were annotated at the trial level, not the criterion level).
Each trial was initially annotated by clinicians followed a two-person blinded annotation paradigm.
In this study, we mainly focus on automatically classifying seven key exclusions for cancer trials, chosen based on their frequency of occurrence as well as clinical significance.
The selected exclusions (with the abbreviations used throughout this paper) are:
\begin{itemize}[parsep=1pt]
  \item prior malignancy (Prior): exclude patients with a previous cancer history
  \item human immunodeficiency virus (HIV): exclude HIV/AIDS positive patients, including well-controlled HIV
  \item hepatitis B virus (HBV): exclude patients with a history of hepatitis B infection
  \item hepatitis C virus (HCV): exclude patients with a history of hepatitis C infection
  \item psychiatric illness (Psych): exclude patients with a history of a major psychiatric/mental health disorder
  \item substance abuse (Subst): exclude patients with a history of substance abuse, including drug and/or alcohol abuse  \item autoimmune disease (Auto): exclude patients with a chronic autoimmune disease (e.g., lupus, rheumatoid arthritis, scleroderma)
\end{itemize}
These clinically relevant criteria may facilitate the selection of a more homogeneous study population for a given trial. However, some controversy surrounds the application or misapplication of some of these criteria, which may inappropriately generate disparities by excluding specific populations of patients whose co-morbid conditions may not be germane to the intervention assessed in a given trial.

For the 764 trials in the dataset, the eligibility criteria section was obtained from ClinicalTrials.gov for each trial and was further divided into individual criteria.
This was done for two main reasons.
First, most BERT-based models have a maximum token input limit of 512, and the original token length in our data ranged up to 2355.
Therefore, the eligibility criteria descriptions were split into individual criteria (which are almost always separated using numbered or unordered lists) to ensure that each criterion fits within this limit.
Second, this approach helped to remove noisy data from the eligibility criteria descriptions using the keyword approach described below. %The whole eligibility section contains multiple conditions from varying aspects, too many unrelated information confuse the model and make it challenging to identify the specific requirements for the criteria we are interested in. Therefore, we process to each criterion in the eligibility criteria separately for text classification. Oftentimes only one or two criterion convey the necessary information for the specific criteria. As a pre-processing step, we implement a simple text cleaning process, including the removal of extra spaces and new lines. These pre-processing steps help to standardize the text data and remove any unnecessary or irrelevant information that can interfere with the text classification process.

The eligibility section is often, but not always, divided into inclusion criteria (a patient must meet all of these) and exclusion criteria (a patient must meet none of these). Sometimes these are not separated and a single nonspecific list of eligibility criteria are provided, but these are clear from reading the text. Importantly, an exclusion is not necessarily only stated in the exclusion criteria section. Here are some examples:

\vspace{-0.15in}
\begin{itemize}
    \item NCT00095875 [Inclusion Section]: \enquote{No other malignancy within the past 5 years except adequately treated carcinoma in situ of the cervix, basal cell or squamous cell skin cancer, or other cancer curatively treated by surgery alone}
    \item NCT00057876 [Exclusion Section]: \enquote{Malignancy within the past 5 years except nonmelanoma skin cancer, carcinoma in situ of the cervix, or organ-confined prostate cancer (Gleason score no greater than 7)}
  \item NCT00048997 [Non-specific Eligibility Section]: \enquote{No other malignancy within the past 3 years except nonmelanoma skin cancer}  
\end{itemize}
\vspace{-0.15in}
All of these trials excluded patients with prior malignancy within a specific time frame, but using different terms under different section headers. We thus use all criteria in the eligibility criteria section but prepared each criterion with an indicator of the section it came from (\enquote{inclusion}, \enquote{exclusion}, \enquote{eligibility} for non-specific section) to provide context. Table 4 shows the sample text input for the classification model. % need to add sample in put data

%\vspace{-5pt}
\subsection{Keyword Filtering}
\label{section:keyword_filtering}
%\vspace{-5pt}

Our dataset was annotated at the trial level, not the individual criterion level.
However, it is the individual criterion that conveys the semantic constraint of the exclusion, so it would make the most sense to focus the classification at the individual criterion level.
In order to accurately find the specific criterion that conveys the given exclusion condition, we created lists of keywords for all 7 exclusion types.
This allows us to convert the task of binary classification at the trial level to binary classification at the criterion level by only classifying criterion that contain one of the selected keywords.
Our goal in creating the keyword lists was recall: keywords alone are insufficient to classify a criterion according to each of the targeted exclusions, a downstream classifier (described later) will perform the binary classification.
However, if a criterion that specifies a targeted exclusion were not to contain one of the specified keywords, then the downstream classifier takes a hit in terms of recall.
Table~\ref{table:keywords} provides the list of keywords used for each exclusion.
%Table~\ref{table:keyword_performance} shows the performance of each exclusion's keywords list.
%As can be seen in the table, the keywords from Table~\ref{table:keywords} 

\begin{table}[h!]
\centering
\caption{Keywords for each criteria}
\vspace{-0.1in}
\label{table:keywords}
\begin{adjustbox}{width=\textwidth}
\begin{tabularx}{\linewidth}{|c |>{\arraybackslash}X|} 
\hline
{\bf Criteria} & {\bf Keywords} \\ [0.5ex] 
\hline
Prior & prior malignancy, concurrent malignancy, prior invasive malignancy, other malignancy, known additional malignancy, squamous cell carcinoma, in-situ, cancer, 3 years, 5 years, five years \\ 
\hline
HIV & human immunodeficiency virus, acquired immunodeficiency syndrome, AIDS-defining malignancy, hiv, AIDS-related illness \\
\hline
HBV & hbv, hepatitis\\
\hline
HCV & hcv, hepatitis\\
\hline
Psych & psychosis, depression, psychiatric, psychological, psychologic, nervous, mental illness, mental disease \\
\hline
Subst & ethanol, abuse, alcohol, alcoholism, illicit substance, drug, drugs, medical marijuana, inadequate liver, illicit substance, addictive, substance misuse, cannabinoids, chronic alcoholism \\
\hline
Auto & uncontrolled systemic, autoimmune \\
\hline
\end{tabularx}
\end{adjustbox}
\end{table}

While in theory, it could be problematic to assume any criterion containing one of the above keywords is positive for the exclusion if the trial as a whole is positive for the exclusion, there is a more important limitation to consider.
Another important feature of PROTECTOR1 is that the annotated exclusions were not based only on the clinical trial description.
There were three primary sources the annotators consulted for whether a given exclusion applied to a clinical trial:
(1) the description on ClinicalTrials.gov,
(2) the original clinical trial protocol,
(3) any publications associated with the trial.
Ideally (and ethically), the eligibility criteria across all three of these would be consistent enough that any inclusion/exclusion stated in the protocol or trial would also be present on ClinicalTrials.gov.
In practice, this is not the case, unfortunately, and PROTECTOR1 does not specify the source(s) of information for the exclusion annotation.
Since our method is designed to work only on the description from ClinicalTrials.gov, manual annotation specific to this project became necessary, as described in the next sub-section.
This did, however, at the same time solve the first problem and allow us to have a criterion-specific label for each of the exclusions.

%\vspace{-5pt}
\subsection{Annotation}
%\vspace{-5pt}

For each exclusion type, all criteria in the 764 trials that matched one of the associated keywords was annotated independently by two annotators using a double-blind paradigm (YY and SJ).
The annotation rule for each criterion used the same standard as the original trial-level annotation.
All discrepancies were resolved through discussion and consensus, including the involvement of a subject-matter expert and curator of the PROTECTOR1 database (EL).
A descriptive summary (sample size and annotation agreement) of the annotated dataset is shown in Table~\ref{table:agreement}.
Due to the small number of samples, the annotation proceeded in a single phase such that consistent disagreements (e.g., what level of granularity counts as an autoimmune disease) were not resolved until the end.
This resulted in low agreement in HBV and Auto (the latter of which was also impacted by its low prevalence, and such imbalance skews the $\kappa$ statistic).
However, during reconciliation and consultation with the subject-matter expert, these disagreements were easily clarified, leading to a better gold standard than the agreement numbers suggest.
The other exclusion types, meanwhile, had excellent levels of agreement. Table~\ref{table:examples} shows some examples of annotated criterion for some criteria.

\begin{table}[t]
    \centering
    \vspace{0.1in}
    \caption{Criterion-level annotation summary}
    \vspace{-0.1in}
    \label{table:agreement}
    \begin{tabular}{|c|c|c|c|c|c|c|c|}
        \hline
         & Prior & HIV & HBV & HCV & Psych & Subst & Auto \\
        \hline
        Sample Size & 529 & 200 & 130 & 282 & 281 & 523 & 54 \\
        \hline
        Cohen's $\kappa$ &0.95 &0.74&0.16&0.89& 0.93&0.98&0.22\\
        \hline
        Agreement Accuracy &0.99& 0.96&0.85&0.95&0.97&0.99&0.89\\
        \hline
    \end{tabular}
\end{table}

\begin{table}[ht]
    \centering
    \vspace{0.1in}
    \caption{Examples of annotated criteria}
    \vspace{-0.1in}
    \label{table:examples}
    \begin{tabularx}{\textwidth}{|c|X|c|}
        \hline
         ClinicalTrials.gov ID & \centering{Criterion Text} & Classification \\
        \hline
        NCT00005047 & eligibility: At least 5 years since other prior systemic chemotherapy & 0: Prior not excluded\\
        \hline
        NCT00216060 & exclusion: No prior history of malignancy in the past 5 years with the exception of basal cell and squamous cell carcinoma of the skin&1: Prior excluded\\
        \hline
        NCT00075803 & exclusion: HIV positive & 1: HIV excluded\\
        \hline
        NCT00114101 & inclusion: Patients must not be human immunodeficiency virus (HIV), hepatitis B surface antigen (HBSag), or hepatitis (Hep) C positive & 1: HBV / HCV /HIV excluded\\
        \hline
        NCT00262067 & exclusion: Known brain or other central nervous system (CNS) metastases & 0: Psych not excluded\\
        \hline
        NCT00022516 & eligibility: No psychiatric or addictive disorders that would preclude study & 1: Psych excluded\\
        \hline
    \end{tabularx}
\end{table}

%\vspace{-5pt}
%\vspace{-0.5in}
\subsection{Machine Learning}
%\vspace{-5pt}

We applied six BERT-based models on all exclusions.
Five of the models are pre-existing BERT models, pre-trained on domain-specific corpora:
\vspace{-0.1in}
\begin{enumerate}[parsep=1pt]
    \item {\bf BioBERT}\cite{lee2020biobert}: The original BERT model was further pre-trained on PubMed abstracts and PMC full-text articles.
    \item {\bf ClinicalBERT} \cite{alsentzer2019publicly}: the BioBERT model further pre-trained on MIMIC-III \cite{johnson2016mimic} notes.
    \item {\bf BlueBERT}\cite{peng2019transfer}: The original BERT model was further pre-trained on PubMed abstracts and MIMIC-III clinical notes.
    \item {\bf PubMedBERT}\cite{gu2021domain}: a from-scratch BERT model pre-trained on PubMed abstracts (notably the from-scratch nature allowed for it to use a domain-specific word piece model).
    \item {\bf SciBERT}\cite{beltagy2019scibert}: the original BERT model further pre-trained on 1.14M full-text papers from Semantic Scholar (which mainly focuses on computer science and biomedicine).
\end{enumerate}
\vspace{-0.1in}
Additionally, since no pre-trained model specific to clinical trial descriptions exists, we further pre-trained the ClinicalBERT model using 442,370 eligibility criteria sections from ClinicalTrials.gov. We used a batch size of 64, a maximum sequence length of 512, and a learning rate of 2e-05. The model trained on all available text for 10,000 steps, and the masked language model probability = 0.15. We denote this model as {\bf ClinicalTrialBERT}.
%The ClinicalTrialBERT is available at [URL] for others in the community to use. % TODO: update this sentence if paper is accepted

\vspace{-0.1in}
\subsection{Evaluation}
\label{subsection:evaluation}

Due to the small sample size (as shown in Table~\ref{table:agreement}), we evaluate using 5-fold cross-validation.
To avoid data leakage, the fold splitting was performed at the trial level (as opposed to the criterion level) such that a single trial with multiple matching criteria does not end up in both the training and testing set for any iteration of cross-validation.

We assess the performance of all classification models using precision, recall, and F1 metrics, and calculate their average values across five folds for evaluation. Each metric was evaluated on both the criterion level (how well the model does at predicting each criterion) and at the trial level (how well the model does at predicting each trial, assuming that a single positive criterion means the trial is positive for that exclusion).

%\vspace{-5pt}
%\subsection{Model Evaluation}
%\vspace{-5pt}
% explain more on the table results. 
%\vspace{-5pt}
\section{Results}
%\vspace{-5pt}

\subsection{Keyword Filtering}

% need to add some analysis of each keyword for psy, sub, and autoimmune
Table~\ref{table:keyword_performance} shows the summary of performance metrics for all exclusion types. 
We conducted error analysis for psychiatric illness (Psych), substance abuse (Subst), and autoimmune disease (Auto), as these have lower overall precision.

\begin{table}[t]
    \centering
      %\vspace{0.1in}
    \caption{Keyword performance metrics for all exclusion types}
    \vspace{-0.1in}
    \label{table:keyword_performance}
    \begin{tabular}{|c|c|c|c|c|c|c|c|}
        \hline
         & Prior & HIV & HBV & HCV & Psych & Subst & Auto \\
        \hline
        Precision & 0.87 & 0.90 &0.98 &0.96 &0.68 &0.27 & 0.62\\
        \hline
        Accuracy &0.82 &0.88&0.74 &0.95 &0.67 &0.27 &0.57 \\
        \hline
        Recall & 0.98&0.97&0.98 &1 &0.99 &1 &0.89\\
        \hline
    \end{tabular}
    \vspace{-0.2in}
\end{table}

In trials with psychiatric exclusions, we note the high frequency of keywords ``psychiatric'' ($n$ = 124) and ``nervous'' ($n$ = 96).
The precision for the keyword ``psychiatric'' is 0.92, while the precision for the keyword ``nervous'' is 0.35.
However, removing the keyword ``nervous'' would cause recall to drop from 0.99 to 0.90, so this keyword is left in for the downstream machine learning classifier to disambiguate.
In trials with substance abuse exclusions, the highest frequency keywords are ``drug'' ($n$ = 269) and ``drugs'' ($n$ = 143).
These keywords, however, are liable to be highly confused with the connotation of ``drug'' meaning treatment (since these are phase III clinical trials, they almost all are drug trials).
As such the precision of ``drug'' is only 0.22 and the precision of ``drugs'' is only 0.23.
Removing these would cause the overall recall will drop from 1.00 to 0.92.
Finally, for trials with autoimmune exclusions, we note there are only 36 trials in the dataset that have this exclusion, and the keywords from Table~\ref{table: keywords} are only able to capture 32 of them.
This results in the recall already being lower than 0.9, while the precision is around 0.6.
So any further sacrifices for the sake of precision would result in an unacceptable loss of recall for this stage of the pipeline.
In summary, we believe that our keyword lists are suitable for capturing the criteria with the considered exclusions, with some consideration for precision but the primary focus being on recall.

\subsection{Exclusion Classification}

\begin{table}[t!]
\centering
  \vspace{0.1in}
  \caption{Evaluation Results of BERT-Based Models Across All Criteria}
  \vspace{-0.1in}
  \label{table:results}
  \small
  \begin{tabular}{l|ccc|ccc|ccc|ccc}
    %\cline{1-13}
    \multicolumn{1}{c}{} & \multicolumn{6}{c|}{Prior} & \multicolumn{6}{c}{HIV} \\ 
    \multicolumn{1}{c}{} & \multicolumn{3}{c}{Criterion Level} & \multicolumn{3}{c|}{Trial Level} & \multicolumn{3}{c}{Criterion Level} & \multicolumn{3}{c}{Trial Level} \\ \hline
     & P & R & F1 & P & R & F1 & P & R & F1 & P & R & F1 \\
    BioBERT &1.00&0.98&\bf{0.99}&1.00&0.98&\bf{0.99}&1.00
    &0.99&\bf{1.00}&1.00&0.99&\bf{1.00}\\
    ClinicalBERT & 0.99&0.99&\bf{0.99}&0.99&0.99&\bf{0.99}&1.00&0.99&\bf{1.00}&1.00&0.99&\bf{1.00}\\
    PubMedBERT & 0.96&0.97&0.96&0.97&0.98&0.97&1.00
    &1.00&\bf{1.00}&1.00&1.00&\bf{1.00}\\
    BlueBERT & 0.98&0.99&\bf{0.99}&0.99&1.00&\bf{0.99}&0.99
    &0.99&0.99&0.99&0.99&0.99\\
    SciBERT & 0.96&0.97&0.97&0.98&0.97&0.97&1.00&1.00&\bf{1.00}&1.00&1.00
    &\bf{1.00}\\
    ClinicalTrialBERT &0.99
    &0.99&{\bf0.99}&0.99&0.99&\bf{0.99}&1.00&1.00&\bf{1.00}&1.00&1.00&\bf{1.00}\\
    \cline{1-13}
     \multicolumn{13}{c}{}\\
    \multicolumn{1}{c}{} & \multicolumn{6}{c|}{Psych} & \multicolumn{6}{c}{HBV} \\ 
    \multicolumn{1}{c}{} & \multicolumn{3}{c}{Criterion Level} & \multicolumn{3}{c|}{Trial Level} & \multicolumn{3}{c}{Criterion Level} & \multicolumn{3}{c}{Trial Level} \\ \hline
     & P & R & F1 & P & R & F1 & P & R & F1 & P & R & F1 \\
    BioBERT &0.99&0.99&0.99&0.99&0.99&0.99&0.99&01.00&0.99&0.99&1.00
    &\bf{1.00}\\

    ClinicalBERT &0.99&1.00&\bf{1.00}&0.99&1.00&\bf{1.00}&0.99&1.00&\bf{1.00}&1.00&1.00
    &\bf{1.00}\\
    PubMedBERT &0.99&1.00&0.99&0.99&1.00&\bf{1.00}&0.98&1.00&0.99&0.99&1.00
    &\bf{1.00}\\
    BlueBERT & 0.98&1.00&0.99&0.99&1.00&0.99&0.99&1.00&0.99&0.99&1.00
    &\bf{1.00}\\

    SciBERT & 0.99&1.00&0.99&0.99&1.00&\bf{1.00}
    &0.99&0.98&0.98&1.00&0.99
    &0.99\\

    ClinicalTrialBERT &0.99&1.00&0.99&0.99&1.00&\bf{1.00}
    &0.98&1.00&0.99&0.99&1.00
    &\bf{1.00}\\
    \cline{1-13}
     \multicolumn{13}{c}{}\\
    \multicolumn{1}{c}{} & \multicolumn{6}{c|}{HCV} & \multicolumn{6}{c}{Auto} \\ 
    \multicolumn{1}{c}{} & \multicolumn{3}{c}{Criterion Level} & \multicolumn{3}{c|}{Trial Level} & \multicolumn{3}{c}{Criterion Level} & \multicolumn{3}{c}{Trial Level} \\ \hline
     & P & R & F1 & P & R & F1 & P & R & F1 & P & R & F1 \\
    BioBERT &0.97&0.98&{\bf0.97}&0.97&0.99&\bf{0.98}&0.98
    &0.98&0.98&0.98&0.98&0.98\\
    ClinicalBERT &0.97&0.96&\bf{0.97}&0.98&0.99&\bf{0.98}&0.98
    &0.98&0.98&0.98&0.98&0.98\\
    PubMedBERT &0.97&0.93&0.94&0.96&0.94&0.95&0.98&1.00&\bf{0.99}&0.98&1.00&\bf{0.99}\\
    BlueBERT &0.94&0.98&0.96&0.96&0.99&0.97&0.96&1.00&0.98&0.96&1.00&0.98\\

    SciBERT &0.96&0.95&0.95&0.96&0.97&0.96&0.98&1.00&\bf{0.99}&0.98&1.00&\bf{0.99}\\

    ClinicalTrialBERT &0.97&0.97&\bf{0.97}&0.98&0.98&\bf{0.98}&0.98&0.98&0.98&0.98&0.98&0.98\\
    \hline
    \multicolumn{13}{c}{}\\
    \multicolumn{1}{c}{} & \multicolumn{6}{c}{Subst}\\ 
    \multicolumn{1}{c}{} & \multicolumn{3}{c}{Criterion Level} & 
    \multicolumn{3}{c}{Trial Level} \\ \cline{1-7}
     & P & R & F1 & P & R & F1 \\

    BioBERT &1.00&1.00&\bf{1.00}&1.00&1.00&{\bf1.00}\\
    ClinicalBERT &0.97&1.00&0.98&0.99&1.00&0.99\\
    PubMedBERT &0.99&1.00&0.99&1.00&1.00&{\bf1.00}\\
    BlueBERT &0.98&1.00&0.99&1.00&1.00&{\bf1.00}\\

    SciBERT &0.98&1.00&0.99&1.00&1.00&{\bf1.00}\\

    ClinicalTrialBERT &0.98&1.00&0.99&1.00&1.00&{\bf1.00}\\

    \end{tabular}
\end{table}

Table~\ref{table: results} presents the results for all six BERT-based models evaluated on the seven exclusion types, including both criterion-level and trial-level assessment.
Across all models, the overall F1 score for all criteria ranged from 0.94 to 1.0 for both evaluation levels. 
For the Prior exclusion, BioBERT, ClinicalBERT, BlueBERT, and the pre-trained ClinicalTrialBERT model collectively attained an F1 score of 0.99, underscoring their top performance in both criterion-level and trial-level evaluations. 
With the exception of BlueBERT, all other models achieved a perfect F1 score of 1.00 in both criterion-level and trial-level evaluations for HIV exclusion.  
ClinicalBERT attained the highest F1 score of 0.99 on the criterion-level evaluation for Psych. 
Additionally, ClinicalBERT, PubMedBERT, SciBERT, and ClinicalTrialBERT achieved a perfect F1 score of 1.00 on the trial-level evaluation.
For the HBV exclusion, ClinicalBERT achieved the highest F1 score of 1.00. 
Except for SciBERT, all other models also attained the highest F1 score of 1.00. Regarding HCV, BioBERT, ClinicalBERT, and ClinicalTrialBERT achieved the highest F1 score of 0.97 at the criterion level, and these three models likewise achieved the highest F1 score of 0.98 during the trial-level evaluation.  
For the autoimmune disease exclusion, both PubMedBERT and SciBERT achieved the highest F1 score of 0.98 on the criterion-level evaluation, and these two models also attained the highest F1 score of 0.99 on the trial-level evaluation.
The BioBERT model demonstrates superior performance over other models when evaluated at the criterion level, achieving an impressive F1 score of 1.00. 
In the context of trial-level evaluation, it is worth noting that ClinicalBERT is the sole exception, as it also achieved an F1 score of 1.00, outperforming all other models in this specific evaluation.

Our results indicate that, at the criterion-level evaluation, there are no differences among all models for HBV and Auto.
Our pre-trained ClinicalTrialBERT model performed equally or better than other models for three out of five remaining exclusions (Prior, HIV, HCV), and performed comparably well against the best-performing model on the other two (0.96 vs. 0.98 on HIV, 0.95 vs. 0.96 on Psych).
At the trial-level evaluation, our pre-trained ClinicalTrialBERT model performed equally or better than other models in two out of five criteria compared to other models (HCV and Subst), while also performing comparably well against the best-performing model on the other three (0.91 vs. 0.93 on Prior, 0.96 vs. 0.98 on HIV, 0.96 vs. 0.97 on Psych).
Meanwhile, BlueBERT and SciBERT had several exclusions for which they performed particularly below the best-performing model. This is to be somewhat expected for SciBERT, which has additional non-biomedical training data.

Five models out of six achieved 1.00 in the F1 score for HIV exclusion, and five models out of six achieved 0.99 in the F1 score for Psych. The reason behind such perfect performance is likely due to the nature of these tasks. HIV and Psych criteria are straightforward, e.g., ``Exclusion: HIV positive'', ``Exclusion: Psychiatric illness''.
The same situation happened in Auto as well.
The sample size for autoimmune is only 54, with only 7 negatives.
Yet the keyword-only precision for Auto is just 0.62, so the model is learning something beyond the keywords. In examining the data, in many cases, this exclusion is simply stated (e.g., ``no autoimmune disease'').
Those are fairly simple criteria, then, compared with descriptions for many of the other exclusions. 

Of all the criteria, HCV yields the poorest performance from all models.
We conducted an error analysis on SciBERT for HCV since it performed the worst overall model performance.
We found the model conflated HCV and HBV (hepatitis C and B viruses, respectively), even though they oftentimes show concurrently in the same sentence.
However, some sentences only mentioned HBV without saying HCV and the model wrongly recognized these as HCV.

\section{Discussion}
%\vspace{-5pt}

Our study demonstrates that using an automatic tool to classify key exclusions from clinical trial eligibility criteria description holds immense potential.
The criterion-level evaluation provides insight into our model's overall performance, while the trial-level evaluation provides a more practical and informative end-user view as it gives a sense of how many trials will be missed or falsely recommended based on each exclusion. 
With this in mind, the best-performing criterion-level model ranged from 0.94 (HCV) to 1.0 (HIV, Psych, Subst) while the best trial-level model ranged from 0.95 (HCV) to 1.00 (HIV, Psych, HBV, Subst).
These results are more than sufficient to enable the scaling up of the types of analyses performed on the 764 PROTECTOR1 trials\cite{pasalic2020association,patel2020exclusion,patel2021transparency} to much larger subsets of the ~445,000 trials currently available from ClinicalTrials.gov.

Our results also suggest that such methods can be used as part of patient-trial matching methods since a large percentage of eligibility criteria are shared across many trials (especially within specific domains such as cancer).
Many eligibility criteria are not applicable to this approach, instead requiring more fine-grained information extraction techniques such as those done by the Criteria2Query system\cite{yuan2019criteria2query}.
Future work in the space of patient-trial matching then should focus on hybrid solutions: differentiating between criteria that require specific facts extracted (and automatically covered to structured queries) and those criteria that are semantically common yet lexically diverse.
The latter type may be better approached using text classification based on the large language models used here since modern language models are excellent at identifying paraphrase-like similarity between sentences that share few words in common.
These types of criteria further do not generally have specific argument structures (e.g., substance abuse criteria do not specifically detail the exact type of substance, its regularity of use, or the exact length of use).
Such criteria are, rather, loose descriptions of common patient features that will be easily recognized and differentiated by clinicians.
Therefore for these kinds of criteria, the approach taken in this work is highly appropriate, so future work should concentrate on differentiating which criteria are of this type and, those, which have the critical mass of frequency to approach with the types of methods studied here.

Prior to this study, to the best of our knowledge, there had been no pre-trained large language model on clinical trial descriptions.
By pre-training such a model on the eligibility criteria section (the main section targeted by NLP systems) from hundreds of thousands of trials, our  ClinicalTrialBERT model will be useful for clinical trial NLP tasks well beyond the current work.
The specific results in this study indicate that ClinicalTrialBERT is a robust model for text classification for clinical trials.
This is not a surprising conclusion, but demonstrating the efficacy of the model across seven tasks is empirically important as we plan to share this BERT model and use it for future clinical trial NLP tasks.

In our study, the original PROTECTOR1 trial level annotation was conducted based on various sources, including ClinicalTrials.gov, available study protocols, and publications.
Another interesting finding is that many trials did not disclose certain key criteria on ClinicalTrials.gov, but only mentioned key exclusion criteria in protocols or publications, which will be investigated in future studies.
Further work includes extending the current text classification framework to include information from protocols or publications. 
This would also allow for identifying which trials publish inconsistent information in various sources, which will ultimately help to improve the quality of clinical trial information provided to the public on ClinicalTrials.gov.

%Despite our model did not show a overwhelming advantage over other domain-specific models, our pre-trained model still achieved the highest F1 score among all criteria. These results indicate that ClinicalTrialBERT is a robust model for text classification for clinical trials. Such result demonstrate the efficacy of the model across seven tasks is empirically important as we plan to share this BERT model and apply it for future clinical trial NLP tasks. However, it is important to note that the pattern for each criteria description is highly similar, which cause the results more predictable. Future studies could consider incorporating more diverse patterns to better evaluate the model's performance in a wider range of scenarios.

\vspace{-0.1in}
\paragraph{Limitations}
Our study is limited by the size of its samples, the number of exclusions considered, and its focus on a specific subset of trials (that is, phase III cancer clinical trials).
First, in terms of sample size, most of the exclusions had only a few hundred annotations.
The high recall of the keywords ensured that the annotations were largely complete, but still, such samples may not generalize well when one goes beyond the scope of trials in this work (see third limitation).
It certainly could mean that the very high performance of our models (with the best-performing F1 ranging from 0.94 to 1.0) is likely overly positive.
Second, practical considerations limited this work to just seven exclusions. PROTECTOR1 has many more exclusions (and inclusions) annotated, so we hope to overcome this limitation with follow-up studies that expand the considered criteria.
Finally, this work was limited to phase III cancer trials (the scope of PROTECTOR1).
Hardly any design considerations of the system are overly specific to either phase III trials or cancer trials, but this scope does limit our ability to generalize the results with high confidence.
Notably, oncology trial specialists are familiar with how other oncology trials are described (and likewise less familiar with trials for other fields [such as cardiology]), so as a result the way patient characteristics are specified for cancer trials may be different than how the same basic information is specified for trials in other diseases or specialties.
We do note, however, that cancer trials make up a substantial portion of all clinical trials, so there certainly is optimism that these methods will be applicable to non-oncology trials.

\vspace{-5pt}
\section{Conclusion}
\vspace{-5pt}
In conclusion, we have successfully trained automatic classifiers using domain-specific BERT-based models to identify seven different exclusion criteria in clinical trials. We conducted evaluations at both the trial and criterion levels to assess the performance of all models.
These evaluations demonstrate the ability of BERT-based models in general, and our new pre-trained ClinicalTrialBERT model in particular, to identify these seven exclusions with high performance in precision, recall, and F1.
Our immediate future plan is to create a more comprehensive and mature model that can identify a significant number of desired criteria.

\vspace{-0.1in}
\paragraph{Acknowledgement}
The research reported in this paper was supported by the National Library of Medicine (NLM) of the National Institutes of Health under award number R01LM011934. 

\footnotesize
\bibliographystyle{vancouver}
\bibliography{References}

\end{document}